\pdfoutput=1

\documentclass[11pt]{article}

\usepackage[final]{acl}

\usepackage{times}
\usepackage{latexsym}
\usepackage{hyperref}

\usepackage[T1]{fontenc}

\usepackage[utf8]{inputenc}

\usepackage{microtype}

\usepackage{inconsolata}

\usepackage{graphicx}
\usepackage{graphicx}  

\usepackage{cuted}
\usepackage{amsmath}
\usepackage{amsfonts}
\usepackage{listings}
\usepackage{xcolor}

\usepackage{algorithm}
\usepackage{algorithmic}
\usepackage{multirow}
\usepackage{graphicx}

\usepackage{amsmath}
\usepackage{multirow}
\usepackage{graphicx}

\usepackage{booktabs}
\usepackage{multirow}

\definecolor{keyword}{rgb}{0,0,0}
\definecolor{comment}{rgb}{0,0,0}
\definecolor{string}{rgb}{0,0,0}
\usepackage{bbding}

\lstdefinestyle{mystyle}{
    backgroundcolor=\color{white},
    commentstyle=\color{comment},
    keywordstyle=\color{keyword},
    numberstyle=\tiny\color{gray},
    stringstyle=\color{string},
    basicstyle=\ttfamily\footnotesize,
    breaklines=true,
    captionpos=b,
    numbers=left,
    numbersep=5pt,
}

\usepackage{newfloat}
\usepackage{listings}
\DeclareCaptionStyle{ruled}{labelfont=normalfont,labelsep=colon,strut=off} 
\floatstyle{ruled}
\newfloat{listing}{tb}{lst}{}
\floatname{listing}{Listing}
\title{\textit{Do It Yourself (DIY)}: Modifying Images for Poems in a Zero-Shot Setting Using Weighted Prompt Manipulation}

\author{
Sofia Jamil{$^1$} \enspace Kotla Sai Charan{$^1$} \enspace \textbf{ Sriparna Saha{$^1$}} \enspace \textbf{ Koustava Goswami{$^2$}} \enspace 
\textbf{ K J Joseph{$^2$}} \\
{{$^1$} {Department of Computer Science \& Engineering, Indian Institute of Technology Patna, India}}  \quad\\
{{$^2$} {Adobe Research} } \\
$^1$\{sofia\_2321cs16, kotla\_2101mc27, sriparna\}@iitp.ac.in \\
$^2$\{koustavag, josephkj\}@adobe.com
}

\begin{document}
\maketitle

\begin{abstract}
Poetry is an expressive form of art that invites multiple interpretations, as readers often bring their own emotions, experiences, and cultural backgrounds into their understanding of a poem. Recognizing this, we aim to generate images for poems and improve these images in a zero-shot setting, enabling audiences to modify images as per their requirements. To achieve this, we introduce a novel \textit{Weighted Prompt Manipulation (WPM)} technique, which systematically modifies attention weights and text embeddings within diffusion models. By dynamically adjusting the importance of specific words, \textit{WPM} enhances or suppresses their influence in the final generated image, leading to semantically richer and more contextually accurate visualizations.  Our approach exploits diffusion models and large language models (LLMs) such as GPT in conjunction with existing poetry datasets, ensuring a comprehensive and structured methodology for improved image generation in the literary domain. To the best of our knowledge, this is the first attempt at integrating weighted prompt manipulation for enhancing imagery in poetic language.
Resources related to data and codes are available here: \href{https://github.com/SofeeyaJ/Weighted-Prompt-Manipulation}{DIY}
\end{abstract}

\section{Introduction}

\begin{figure}[!htbp]
\centerline{\includegraphics[width=\columnwidth]{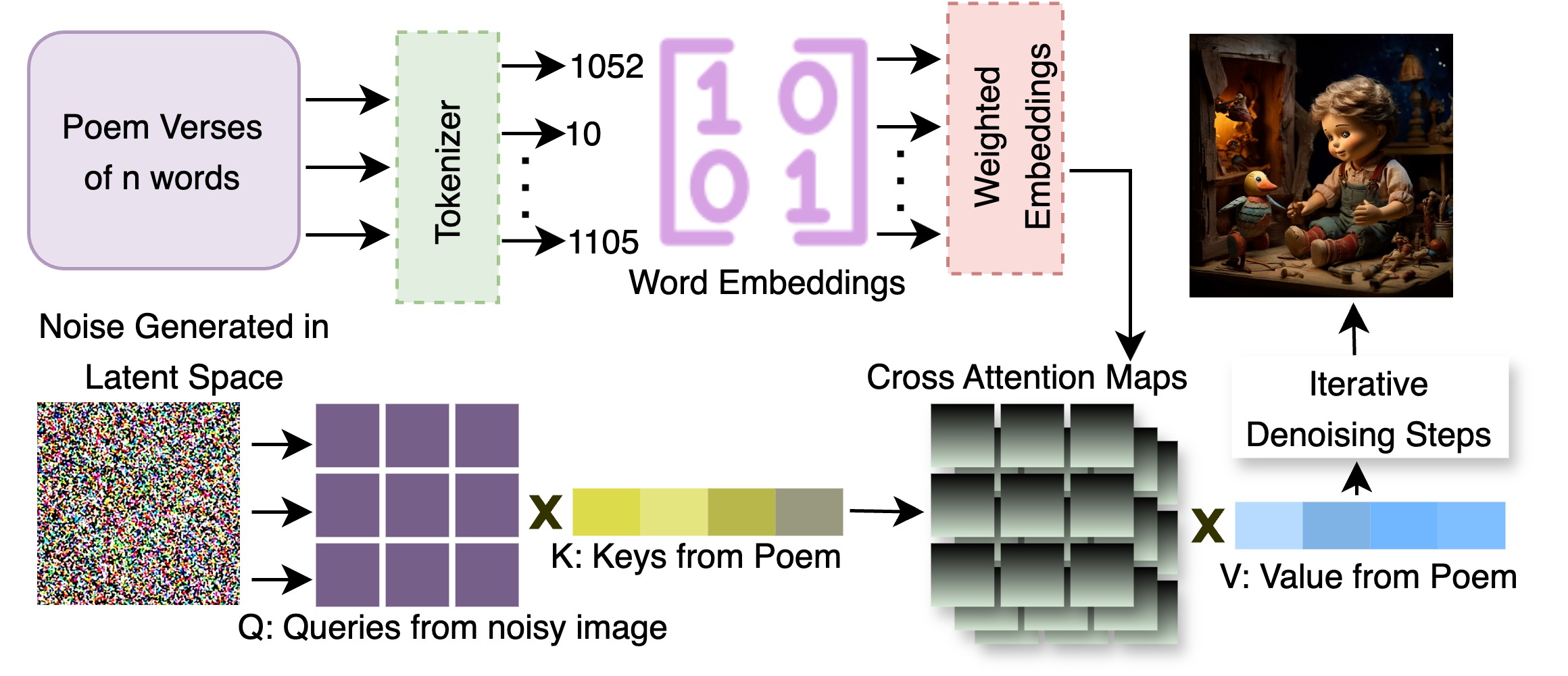}}
\caption{Architectural diagram of the poem-to-image generation process using our proposed \textit{Weighted Prompt Manipulation} technique.}
\label{modelarchitecture}
\end{figure}

Recent advancements in diffusion models have transformed the landscape of generative AI. These text to image generation models are pretrained on vast datasets of image-text pairs \cite{laion_1,laion_2}, and leverage state-of-the-art techniques, including large-scale pre-trained language models \cite{pretrained,pretrained_2,pretrained_3}, variational autoencoders \cite{variational_autoencoeders}, and diffusion-based architectures \cite{dalle,stable_diffusion}. As a result, they excel in generating highly realistic and visually compelling images. 
However, current diffusion models often struggle to interpret metaphorical language, symbolism, and nuanced themes. Therefore, creative fields like poetry fail to directly generate relevant visuals and often lead to inconsistent or inaccurate visual outputs.
To address this limitation, we propose \textbf{Weighted Prompt Manipulation}, a novel approach illustrated in Figure \ref{modelarchitecture}, designed to refine generated images in a real-time setting and adjust their alignment, especially for poetic content. Existing text-to-image editing techniques \cite{editing_1,editing_2,editing_3} have demonstrated remarkable success in tasks such as image translation, style transfer, and appearance modification, all while preserving structural integrity and scene composition. Among these methods, attention layers play a pivotal role in regulating image layout and ensuring a coherent relationship between the generated image and its textual prompt. However, these techniques have not yet been applied in the domain of poems. Therefore, motivated by this, we specifically investigate the attribution of image generation in diffusion models, posing a fundamental question: \textit{How do diffusion models generate images for poems?} To explore this, we employ prompt tuning and a systematic analysis of attention map generation, providing deeper insights into the underlying mechanisms of poem to image synthesis using diffusion-based models. Building on our findings, we introduce \textit{Weighted Prompt Manipulation}, a technique designed to enhance image generation for poetic inputs by improving relevance and fidelity.
Our key contributions include:
\newline \textbf{1.} We introduce a new task of poem visualization, focusing on generating images that accurately capture the rich and intricate details conveyed in poetic text.
\newline \textbf{2.} We propose a training-free \textbf{Weighted Prompt Manipulation} approach, which manipulates images by dynamically adjusting word importance in a real-time setting.
\newline \textbf{3.} We provide a detailed analysis of text-to-image generation within diffusion models, leveraging heat maps and attention maps to better understand how different parts of a poem influence image generation.
\newline \textbf{4.} We conduct extensive quantitative and human evaluations to demonstrate that the diffusion model can be manipulated to enhance image generation by selectively reinforcing specific textual elements without significantly altering the existing visual composition.



\section{Background and Related Works}

Recent advancements in text-driven image manipulation have been significantly influenced by GANs \cite{gan_1,gan_2,gan_3} combined with image-text representations like CLIP \cite{clip}. These approaches enable realistic image modifications using textual input \cite{__21,__7,_43, DBLP:conf/aaai/GoswamiKUJS24,DBLP:conf/iccv/AgarwalKJSGS23} However, while they perform well in structured domains (e.g., human face editing), they often struggle with diverse datasets where subjects vary significantly. To address this, fine-tuning methods \cite{_fine_1,_fine_2,_fine_4} allow models to learn novel styles from just a few images. However, these methods are prone to overfitting, leading to image degradation or content leakage. Alternative approaches, such as Textual Inversion \cite{textualinversion} and Hard Prompt Made Easy (PEZ) \cite{hardpromptmadeeasy}, aim to find optimal text representations (e.g., embeddings or tokens) that capture an object’s characteristics without modifying the underlying text-to-image model parameters. Another line of research focuses on encoder-based methods \cite{_encoder_1,_encoder_2,_encoder_3,_encoder_4}, which use visual encoders to extract image features and map them to text prompts. While these methods have set the standard in state-of-the-art performance, they remain limited by the capabilities of visual encoders, which often struggle with capturing fine-grained textures beyond abstract style information.

\begin{figure}[!ht]
\centerline{\includegraphics[width=\columnwidth]{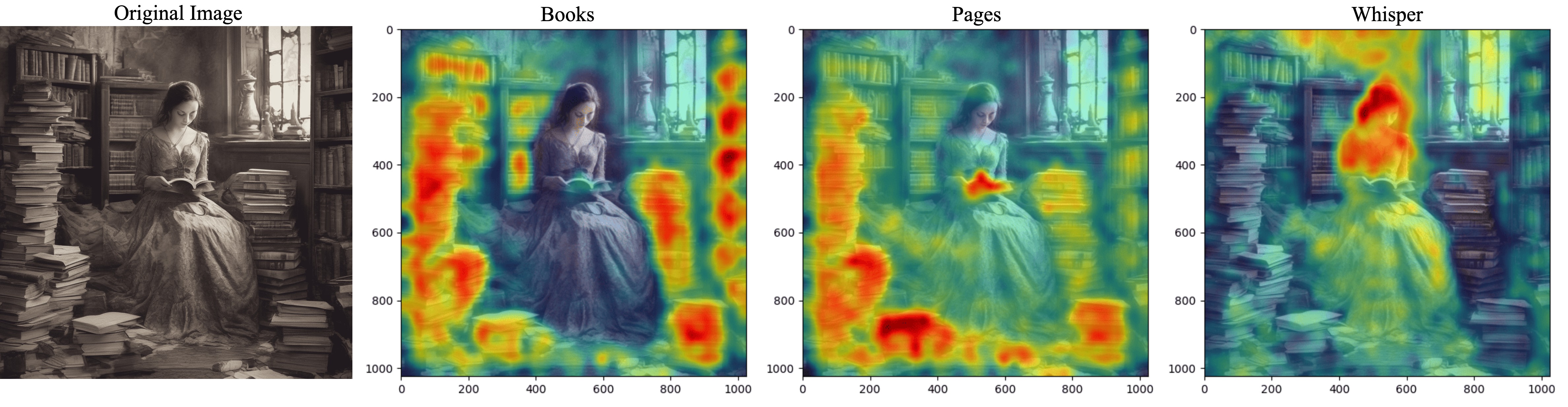}}
\caption{Heat maps for the generated image highlighting captured and missed words from the prompt. Readers are encouraged to zoom in for improved visibility.}
\label{heatmaps}
\end{figure}

\section{Tasks Setups}

\begin{figure}[!ht]
\centerline{\includegraphics[width=\columnwidth]{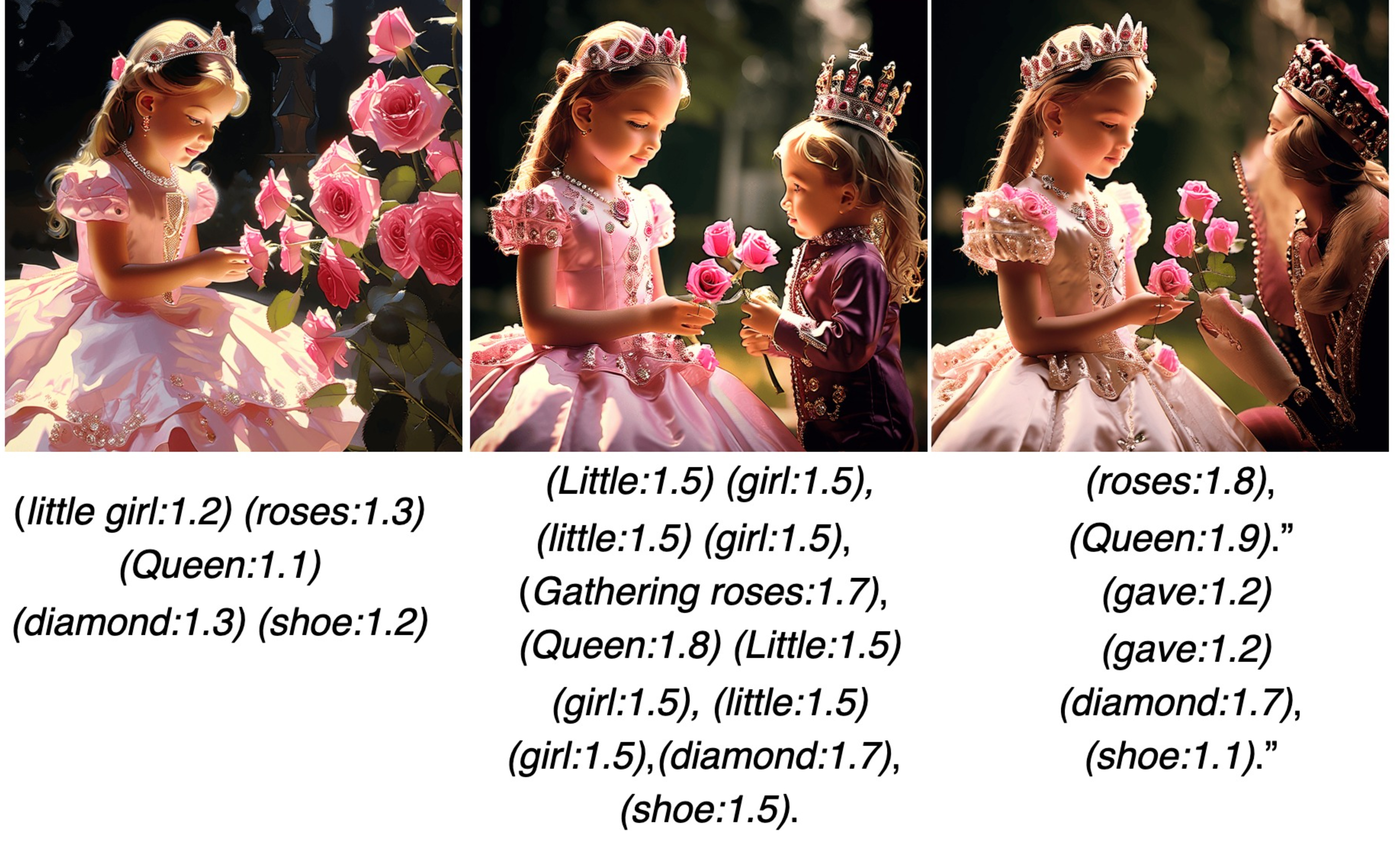}}
\caption{Examples of images generated using various weighted prompts, with corresponding weights displayed below each image.}
\label{wpm}
\end{figure}

\textbf{Challenges in the Poem to Image
generation: }
\newline Building on the efficacy of the Playground \cite{playgroundv3} diffusion model in image generation, we conduct an in-depth analysis of how diffusion models process different words in poetry \cite{poemtale, coling_paper}. As illustrated in Figure \ref{heatmaps}, diffusion models exhibit a strong bias toward visual elements, with the highest attention given to concrete objects (`books', `page'). Diffusion models leverage CLIP embeddings, which are inherently designed to align textual descriptions with corresponding visual features. As a result, CLIP embeddings emphasize words containing visual objects, as they provide explicit semantic grounding for image synthesis. Additionally, the cross-attention mechanism in diffusion models determines how strongly each word contributes to the generated image. Certain words tend to have higher attention scores, guiding the model’s output more effectively, whereas others, being more contextual than structural, receive lower attention weights and have less impact on the final image.
\newline \textbf{Proposed Solution:}
\newline To address the inherent bias of diffusion models toward certain words and their limited attention to others, we propose \textbf{\textit{Weighted Prompt Manipulation (WPM)}} approach. As demonstrated in Figure \ref{modelarchitecture}, it is a systematic approach to dynamically adjust word influence during image generation. By assigning custom weight values to specific words in the prompt, we can enhance the model’s focus on critical poetic elements, ensuring a more faithful and semantically rich visual representation. In our approach, words that naturally receive high attention are explicitly reinforced, while those that receive lower attention are strategically amplified to balance their contribution. Diffusion models use cross-attention mechanisms to determine the importance of each word in a text prompt. \textit{WPM} modifies the default attention scores by explicitly assigning weights to different words, guiding the model to generate images that more accurately capture the semantic depth and poetic meaning. Each word in the prompt is assigned a scaling factor in parentheses. Words with higher weights are given greater prominence in the generated image, while those with lower weights are de-emphasized. As illustrated in Figure \ref{wpm}, the subsequent images are produced using \textit{WPM}. To understand the weighting of certain words that are visually significant in poetry, we employed \texttt{GPT-4o-mini} for image instruction generation.
Our method begins by providing GPT with an initial prompt as demonstrated in Figure \ref{prompot}:
\begin{figure}[!ht]
\centerline{\includegraphics[width=0.9\columnwidth]{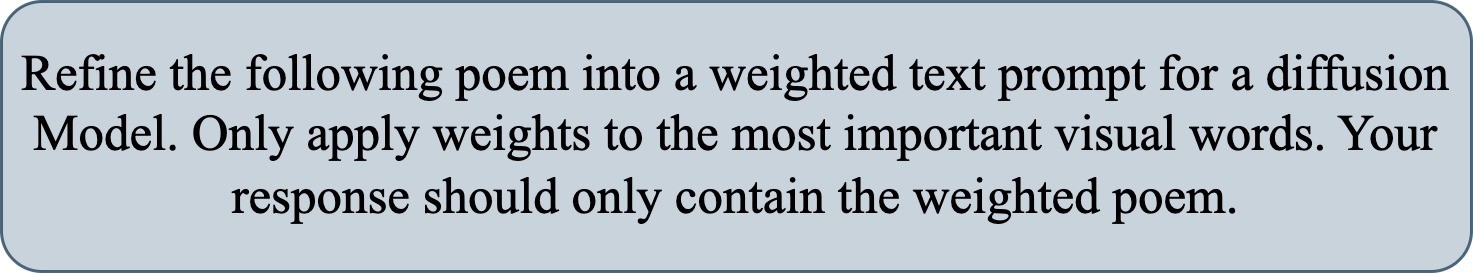}}
\caption{Initial Prompt for WPM}
\label{prompot}
\end{figure}
\newline \textbf{<Input Poem>}
\newline\textit{“Little girl, little girl,
Where have you been?”
\newline“Gathering roses,
To give to the Queen.”
\newline“Little girl, little girl,
What she gave you?
\newline “She gave me diamond,
As big as my shoe.”}
\newline \textbf{<GPT’s Response (Weighted Prompt)>}
\newline \textit{ Little girl, little girl, (girl:1.6)  
Where have you been?”  
\newline“Gathering (roses:1.7),  
To give to the (Queen:1.6).”  
\newline“Little girl, little girl, (girl:1.6)  
What she gave you?  
\newline“She gave me (diamond:1.8),  
As big as my (shoe:1.5).”}

The weighted output generated by GPT is then passed to the diffusion model. \textit{WPM} processes input text by identifying attention markers, such as (word:1.5) to increase emphasis. The corresponding weights are applied to the text embeddings of their respective words and integrated into the model’s dual text encoders. The final weighted embeddings are then used to condition image generation. \textit{(Algorithm \ref{wpmalgorithm} in Appendix contains the entire pseudocode)}.

\begin{figure*}[!htbp]
\centerline{\includegraphics[width=1.7\columnwidth]{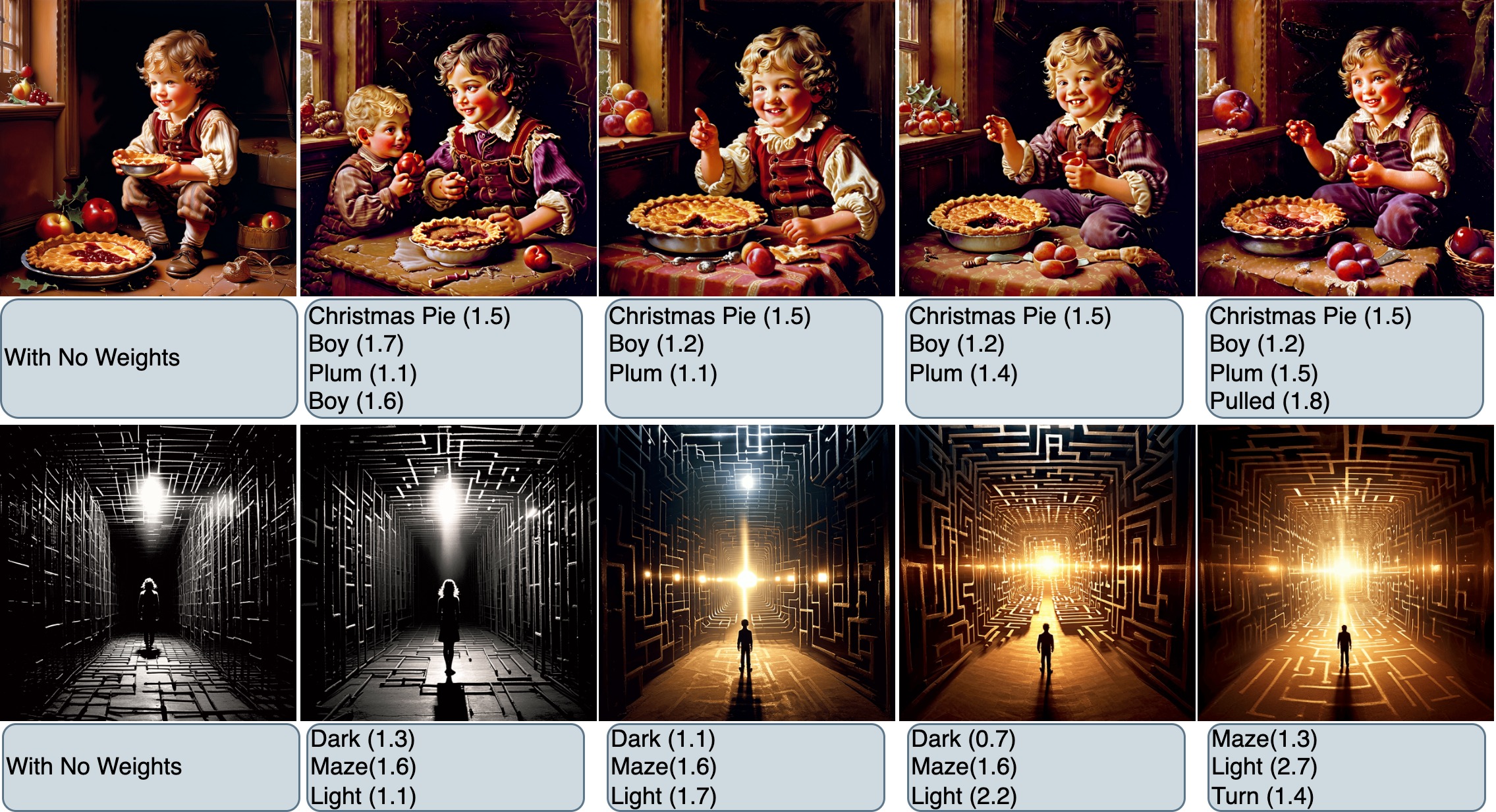}}
\caption{A comparison of generated images using different weights for various words in the same poem. All poems, along with their corresponding weighted prompts specified in the poem, are provided in the Appendix Table \ref{qualitative_poems_results}.}
\label{qualitative}
\end{figure*}

\section{Experiments}

\subsection{Implementation Details}

We implemented our \textit{WPM} approach using three text-to-image models, Playground V3 \cite{playgroundv3}, Stable Diffusion XL \cite{sdxl_refiner}, and Sana \cite{sana}, selected for their training-free, pluggable design, enabling cross-architecture comparisons. To evaluate the alignment between the generated images and poems we employed BLIP \cite{blip} to generate captions for the images and measure their similarity to the original poem. Similarly, we applied Long-CLIP \cite{longclip} to compute the cosine similarity between the poem and the generated image. Experiments were conducted on two benchmark datasets: \textit{PoemSum} \cite{poemsum}, containing 3,011 poems with curated English summaries from Poem Analysis, and \textit{MiniPo} \cite{jamil-etal-2025-poetry}, comprising 1,001 nursery rhymes,  both sourced from online platforms.

\subsection{Results and Discussions}


\subsubsection{Quantitative Evaluation}

To evaluate the effectiveness of our proposed methodology, we present the results in Table \ref{quantitative}. Our \textit{Weighted Prompt Manipulation} approach consistently outperforms direct poem as prompts. Given that the Long-CLIP score measures semantic consistency between text and image, the results demonstrate that incorporating weighted poems into diffusion models yields higher scores, particularly when using the optimal prompt refined through human feedback.
\begin{table}[!htpb]
\centering
\resizebox{\columnwidth}{!}{%
\begin{tabular}{llccccc}
\hline
\textbf{} & \textbf{} & \multicolumn{1}{l}{\textbf{Direct Poem}} & \multicolumn{1}{l}{\textbf{Prompt 1}} & \multicolumn{1}{l}{\textbf{Prompt 2}} & \multicolumn{1}{l}{\textbf{Prompt 3}} & \multicolumn{1}{l}{\textbf{Prompt 4}} \\ \hline
\multirow{3}{*}{\textit{BLIP}} & Stable Diffusion & 0.2243 & 0.2325 & 0.2340 & 0.2412 & 0.2352 \\
 & Playground V3 & 0.3296 & 0.3270 & 0.3408 & 0.3317 & 0.3272 \\
 & Sana & 0.3148 & 0.3365 & 0.3380 & 0.3356 & 0.3354 \\ \hline
\multirow{3}{*}{\textit{LongClip}} & Stable Diffusion & 0.2391 & 0.2245 & 0.2112 & 0.2309 & 0.2273 \\
 & Playground V3 & 0.2480 & 0.2449 & 0.2489 & 0.2507 & 0.2494 \\
 & Sana & 0.2286 & 0.2388 & 0.2387 & 0.2384 & 0.2418 \\ \hline
\end{tabular}%
}
\caption{Quantitative evaluation of generated images using different diffusion models on different prompts.}
\label{quantitative}
\end{table}
Notably, our \textit{WPM} technique is broadly applicable to all Stable Diffusion style models. Experimental results on SD 3.5 Medium and Playground v3 further validate the adaptability of our approach across various diffusion-based models.

\begin{table}[!ht]
\centering
\resizebox{\columnwidth}{!}{%
\begin{tabular}{lcc}
\hline
\textbf{Metrics} & \multicolumn{1}{l}{\textbf{WPM}} & \multicolumn{1}{l}{\textbf{Without WPM(Only Poem)}} \\ \hline
\textit{Meaning} & 4.3 & 3.8 \\
\textit{Visual Objects (Nouns)} & 4.2 & 4.35 \\
\textit{Image Aesthetics} & 3.1 & 3 \\
\textit{Action Depicted (Verbs)} & 3.9 & 3.2 \\ \hline
\end{tabular}%
}
\caption{The results of Human Evaluation Scores in terms of expert ratings (1-5).}
\label{humanevaluations}
\end{table}

\subsubsection{Human Evaluation}

Given that existing automated metrics may not fully capture the quality of generated images and with no standardized metric available, we incorporated human evaluations. We selected 5\% of the samples from the \textit{PoemSum} dataset and had domain experts review images generated both with and without our \textit{WPM} approach. Each image was evaluated based on four key criteria: interpretability of meaning, visual objects, image aesthetics, and action depicted. Participants rated each sample on a scale of 1 to 5, with higher scores indicating better quality. The final rating for each image was determined by averaging the scores provided by three experts. To ensure unbiased assessments, the evaluators were not informed of the model used to generate each image. As shown in Table \ref{humanevaluations}, the results demonstrate that \textit{WPM} significantly improves image generation in terms of semantic meaning and alignment. Moreover, we conducted qualitative evaluations to compare the results of \textit{Weighted Prompt Manipulation} with those generated without it. Our observations indicate that images produced using weighted prompts are able to incorporate certain key elements that were otherwise missing when plain poems were used as prompts. As illustrated in Figure \ref{qualitative}, when the diffusion model processes only the raw poem, the generated images tend to emphasize specific words \textit{(pie, landscape, maze)} while completely ignoring others \textit{(plum, smoke, light)}. However, by assigning greater importance to the previously ignored words, the updated images successfully incorporate those elements alongside the already emphasized ones.


\section{Conclusion}
In this work, we propose the task of poem-to-image manipulation based on the reader's interpretation in a zero-shot setting. Our novel \textit{Weighted Prompt Manipulation} technique systematically modifies attention weights and text embeddings within diffusion models to add or remove certain elements in the poem-to-image generation. To evaluate the effectiveness of our method, we conduct extensive experiments on benchmark poetry visualization datasets. Our evaluation framework includes human assessments, qualitative analyses, and quantitative metrics, ensuring a comprehensive assessment of our approach. In future work, we aim to apply consistent weighted attention to phrases instead of individual words, making it a scalable poetry visualization tool that enables real-world applications in education, cultural preservation, and literary content creation.

\section{Limitation}

A key limitation of our \textit{Weighted Prompt Manipulation (WPM)} approach is its effectiveness in handling poems that lack explicit visual elements or rely heavily on abstract concepts. Since our method primarily enhances image generation by adjusting prompt weights based on the presence of tangible objects and discernible themes, it struggles with highly conceptual or non-visual poetry. In such cases, where the essence of the poem cannot be easily translated into concrete imagery, \textit{WPM} fails to introduce significant variations in the generated outputs. As a result, the images produced remain largely similar across different prompts, limiting the impact of our approach in capturing the deeper, non-representational meanings of such poems.

\section{Ethical Consideration}

A key ethical consideration involves the inherent biases present in diffusion models, which may reflect societal, cultural, or data-driven biases from the pre-trained models. These biases can potentially influence the generation of images related to poems on specific topics or forms, resulting in unfair or inappropriate outputs. To ensure compliance and ethical integrity, we also obtained formal approval from our institute's ethical review board (ERB) before utilizing the dataset and models for research purposes. 

\section{Acknowledgement}
Sriparna Saha would like to acknowledge the funding from ADOBE Research for conducting this research.

\bibliography{acl_latex}

\begin{thebibliography}{38}
\providecommand{\natexlab}[1]{#1}

\bibitem[{Abdal et~al.(2021)Abdal, Zhu, Mitra, and Wonka}]{editing_1}
Rameen Abdal, Peihao Zhu, Niloy~J Mitra, and Peter Wonka. 2021.
\newblock Styleflow: Attribute-conditioned exploration of stylegan-generated images using conditional continuous normalizing flows.
\newblock \emph{ACM Transactions on Graphics (ToG)}, 40(3):1--21.

\bibitem[{Agarwal et~al.(2023)Agarwal, Karanam, Joseph, Saxena, Goswami, and Srinivasan}]{DBLP:conf/iccv/AgarwalKJSGS23}
Aishwarya Agarwal, Srikrishna Karanam, K.~J. Joseph, Apoorv Saxena, Koustava Goswami, and Balaji~Vasan Srinivasan. 2023.
\newblock \href {https://doi.org/10.1109/ICCV51070.2023.00217} {{A-STAR:} test-time attention segregation and retention for text-to-image synthesis}.
\newblock In \emph{{IEEE/CVF} International Conference on Computer Vision, {ICCV} 2023, Paris, France, October 1-6, 2023}, pages 2283--2293. {IEEE}.

\bibitem[{Ahn et~al.(2024)Ahn, Lee, Lee, Kim, Kim, Nam, and Hong}]{_fine_2}
Namhyuk Ahn, Junsoo Lee, Chunggi Lee, Kunhee Kim, Daesik Kim, Seung-Hun Nam, and Kibeom Hong. 2024.
\newblock Dreamstyler: Paint by style inversion with text-to-image diffusion models.
\newblock In \emph{Proceedings of the AAAI Conference on Artificial Intelligence}, volume~38, pages 674--681.

\bibitem[{Andonian et~al.(2021)Andonian, Osmany, Cui, Park, Jahanian, Torralba, and Bau}]{__7}
Alex Andonian, Sabrina Osmany, Audrey Cui, YeonHwan Park, Ali Jahanian, Antonio Torralba, and David Bau. 2021.
\newblock Paint by word.
\newblock \emph{arXiv preprint arXiv:2103.10951}.

\bibitem[{Bau et~al.(2020)Bau, Strobelt, Peebles, Wulff, Zhou, Zhu, and Torralba}]{editing_2}
David Bau, Hendrik Strobelt, William Peebles, Jonas Wulff, Bolei Zhou, Jun-Yan Zhu, and Antonio Torralba. 2020.
\newblock Semantic photo manipulation with a generative image prior.
\newblock \emph{arXiv preprint arXiv:2005.07727}.

\bibitem[{Brock et~al.(2018)Brock, Donahue, and Simonyan}]{gan_1}
Andrew Brock, Jeff Donahue, and Karen Simonyan. 2018.
\newblock Large scale gan training for high fidelity natural image synthesis.
\newblock \emph{arXiv preprint arXiv:1809.11096}.

\bibitem[{Brown et~al.(2020)Brown, Mann, Ryder, Subbiah, Kaplan, Dhariwal, Neelakantan, Shyam, Sastry, Askell et~al.}]{pretrained_3}
Tom Brown, Benjamin Mann, Nick Ryder, Melanie Subbiah, Jared~D Kaplan, Prafulla Dhariwal, Arvind Neelakantan, Pranav Shyam, Girish Sastry, Amanda Askell, et~al. 2020.
\newblock Language models are few-shot learners.
\newblock \emph{Advances in neural information processing systems}, 33:1877--1901.

\bibitem[{Chen et~al.(2023)Chen, Pan, Yao, and Mei}]{_encoder_1}
Jingwen Chen, Yingwei Pan, Ting Yao, and Tao Mei. 2023.
\newblock Controlstyle: Text-driven stylized image generation using diffusion priors.
\newblock In \emph{Proceedings of the 31st ACM International Conference on Multimedia}, pages 7540--7548.

\bibitem[{Devlin et~al.(2019)Devlin, Chang, Lee, and Toutanova}]{pretrained}
Jacob Devlin, Ming-Wei Chang, Kenton Lee, and Kristina Toutanova. 2019.
\newblock Bert: Pre-training of deep bidirectional transformers for language understanding.
\newblock In \emph{Proceedings of the 2019 conference of the North American chapter of the association for computational linguistics: human language technologies, volume 1 (long and short papers)}, pages 4171--4186.

\bibitem[{Frenkel et~al.(2024)Frenkel, Vinker, Shamir, and Cohen-Or}]{_fine_4}
Yarden Frenkel, Yael Vinker, Ariel Shamir, and Daniel Cohen-Or. 2024.
\newblock Implicit style-content separation using b-lora.
\newblock In \emph{European Conference on Computer Vision}, pages 181--198. Springer.

\bibitem[{Gal et~al.(2022{\natexlab{a}})Gal, Alaluf, Atzmon, Patashnik, Bermano, Chechik, and Cohen-Or}]{textualinversion}
Rinon Gal, Yuval Alaluf, Yuval Atzmon, Or~Patashnik, Amit~H Bermano, Gal Chechik, and Daniel Cohen-Or. 2022{\natexlab{a}}.
\newblock An image is worth one word: Personalizing text-to-image generation using textual inversion.
\newblock \emph{arXiv preprint arXiv:2208.01618}.

\bibitem[{Gal et~al.(2022{\natexlab{b}})Gal, Patashnik, Maron, Bermano, Chechik, and Cohen-Or}]{__21}
Rinon Gal, Or~Patashnik, Haggai Maron, Amit~H Bermano, Gal Chechik, and Daniel Cohen-Or. 2022{\natexlab{b}}.
\newblock Stylegan-nada: Clip-guided domain adaptation of image generators.
\newblock \emph{ACM Transactions on Graphics (TOG)}, 41(4):1--13.

\bibitem[{Gao et~al.(2024)Gao, Liu, Sun, Tang, Zeng, Chen, and Zhao}]{_encoder_2}
Junyao Gao, Yanchen Liu, Yanan Sun, Yinhao Tang, Yanhong Zeng, Kai Chen, and Cairong Zhao. 2024.
\newblock Styleshot: A snapshot on any style.
\newblock \emph{arXiv preprint arXiv:2407.01414}.

\bibitem[{Goswami et~al.(2024)Goswami, Karanam, Udhayanan, Joseph, and Srinivasan}]{DBLP:conf/aaai/GoswamiKUJS24}
Koustava Goswami, Srikrishna Karanam, Prateksha Udhayanan, K.~J. Joseph, and Balaji~Vasan Srinivasan. 2024.
\newblock \href {https://doi.org/10.1609/AAAI.V38I16.29766} {Copl: Contextual prompt learning for vision-language understanding}.
\newblock In \emph{Thirty-Eighth {AAAI} Conference on Artificial Intelligence, {AAAI} 2024, Thirty-Sixth Conference on Innovative Applications of Artificial Intelligence, {IAAI} 2024, Fourteenth Symposium on Educational Advances in Artificial Intelligence, {EAAI} 2014, February 20-27, 2024, Vancouver, Canada}, pages 18090--18098. {AAAI} Press.

\bibitem[{Houlsby et~al.(2019)Houlsby, Giurgiu, Jastrzebski, Morrone, De~Laroussilhe, Gesmundo, Attariyan, and Gelly}]{_fine_1}
Neil Houlsby, Andrei Giurgiu, Stanislaw Jastrzebski, Bruna Morrone, Quentin De~Laroussilhe, Andrea Gesmundo, Mona Attariyan, and Sylvain Gelly. 2019.
\newblock Parameter-efficient transfer learning for nlp.
\newblock In \emph{International conference on machine learning}, pages 2790--2799. PMLR.

\bibitem[{Jamil et~al.(2025{\natexlab{a}})Jamil, Reddy, Kumar, Saha, Goswami, and Joseph}]{poemtale}
Sofia Jamil, Bollampalli~Areen Reddy, Raghvendra Kumar, Sriparna Saha, Koustava Goswami, and K.~J. Joseph. 2025{\natexlab{a}}.
\newblock \href {https://arxiv.org/abs/2507.13708} {Poemtale diffusion: Minimising information loss in poem to image generation with multi-stage prompt refinement}.
\newblock \emph{Preprint}, arXiv:2507.13708.

\bibitem[{Jamil et~al.(2025{\natexlab{b}})Jamil, Reddy, Kumar, Saha, J, and Goswami}]{jamil-etal-2025-poetry}
Sofia Jamil, Bollampalli~Areen Reddy, Raghvendra Kumar, Sriparna Saha, Joseph~K. J, and Koustava Goswami. 2025{\natexlab{b}}.
\newblock \href {https://aclanthology.org/2025.coling-main.620/} {Poetry in pixels: Prompt tuning for poem image generation via diffusion models}.
\newblock In \emph{Proceedings of the 31st International Conference on Computational Linguistics}, pages 9224--9237, Abu Dhabi, UAE. Association for Computational Linguistics.

\bibitem[{Jamil et~al.(2025{\natexlab{c}})Jamil, Reddy, Kumar, Saha, Joseph, and Goswami}]{coling_paper}
Sofia Jamil, Bollampalli~Areen Reddy, Raghvendra Kumar, Sriparna Saha, K~J Joseph, and Koustava Goswami. 2025{\natexlab{c}}.
\newblock \href {https://arxiv.org/abs/2501.05839} {Poetry in pixels: Prompt tuning for poem image generation via diffusion models}.
\newblock \emph{Preprint}, arXiv:2501.05839.

\bibitem[{Karras et~al.(2021)Karras, Aittala, Laine, H{\"a}rk{\"o}nen, Hellsten, Lehtinen, and Aila}]{gan_2}
Tero Karras, Miika Aittala, Samuli Laine, Erik H{\"a}rk{\"o}nen, Janne Hellsten, Jaakko Lehtinen, and Timo Aila. 2021.
\newblock Alias-free generative adversarial networks.
\newblock \emph{Advances in neural information processing systems}, 34:852--863.

\bibitem[{Karras et~al.(2019)Karras, Laine, and Aila}]{gan_3}
Tero Karras, Samuli Laine, and Timo Aila. 2019.
\newblock A style-based generator architecture for generative adversarial networks.
\newblock In \emph{Proceedings of the IEEE/CVF conference on computer vision and pattern recognition}, pages 4401--4410.

\bibitem[{Kingma et~al.(2019)Kingma, Welling et~al.}]{variational_autoencoeders}
Diederik~P Kingma, Max Welling, et~al. 2019.
\newblock An introduction to variational autoencoders.
\newblock \emph{Foundations and Trends{\textregistered} in Machine Learning}, 12(4):307--392.

\bibitem[{Lang et~al.(2021)Lang, Gandelsman, Yarom, Wald, Elidan, Hassidim, Freeman, Isola, Globerson, Irani et~al.}]{editing_3}
Oran Lang, Yossi Gandelsman, Michal Yarom, Yoav Wald, Gal Elidan, Avinatan Hassidim, William~T Freeman, Phillip Isola, Amir Globerson, Michal Irani, et~al. 2021.
\newblock Explaining in style: training a gan to explain a classifier in stylespace.
\newblock In \emph{Proceedings of the IEEE/CVF International Conference on Computer Vision}, pages 693--702.

\bibitem[{Li et~al.(2022)Li, Li, Xiong, and Hoi}]{blip}
Junnan Li, Dongxu Li, Caiming Xiong, and Steven Hoi. 2022.
\newblock \href {https://arxiv.org/abs/2201.12086} {Blip: Bootstrapping language-image pre-training for unified vision-language understanding and generation}.
\newblock \emph{Preprint}, arXiv:2201.12086.

\bibitem[{Liu et~al.(2024)Liu, Akhgari, Visheratin, Kamko, Xu, Shrirao, Lambert, Souza, Doshi, and Li}]{playgroundv3}
Bingchen Liu, Ehsan Akhgari, Alexander Visheratin, Aleks Kamko, Linmiao Xu, Shivam Shrirao, Chase Lambert, Joao Souza, Suhail Doshi, and Daiqing Li. 2024.
\newblock \href {https://arxiv.org/abs/2409.10695} {Playground v3: Improving text-to-image alignment with deep-fusion large language models}.
\newblock \emph{Preprint}, arXiv:2409.10695.

\bibitem[{Mahbub et~al.(2023)Mahbub, Khan, Anuva, Shahriar, Laskar, and Ahmed}]{poemsum}
Ridwan Mahbub, Ifrad Khan, Samiha Anuva, Md~Shihab Shahriar, Md~Tahmid~Rahman Laskar, and Sabbir Ahmed. 2023.
\newblock \href {https://doi.org/10.18653/v1/2023.emnlp-main.920} {Unveiling the essence of poetry: Introducing a comprehensive dataset and benchmark for poem summarization}.
\newblock In \emph{Proceedings of the 2023 Conference on Empirical Methods in Natural Language Processing}, pages 14878--14886, Singapore. Association for Computational Linguistics.

\bibitem[{Patashnik et~al.(2021)Patashnik, Wu, Shechtman, Cohen-Or, and Lischinski}]{_43}
Or~Patashnik, Zongze Wu, Eli Shechtman, Daniel Cohen-Or, and Dani Lischinski. 2021.
\newblock Styleclip: Text-driven manipulation of stylegan imagery.
\newblock In \emph{Proceedings of the IEEE/CVF international conference on computer vision}, pages 2085--2094.

\bibitem[{Podell et~al.(2024)Podell, English, Lacey, Blattmann, Dockhorn, M{\"u}ller, Penna, and Rombach}]{sdxl_refiner}
Dustin Podell, Zion English, Kyle Lacey, Andreas Blattmann, Tim Dockhorn, Jonas M{\"u}ller, Joe Penna, and Robin Rombach. 2024.
\newblock \href {https://openreview.net/forum?id=di52zR8xgf} {{SDXL}: Improving latent diffusion models for high-resolution image synthesis}.
\newblock In \emph{The Twelfth International Conference on Learning Representations}.

\bibitem[{Radford et~al.(2021)Radford, Kim, Hallacy, Ramesh, Goh, Agarwal, Sastry, Askell, Mishkin, Clark et~al.}]{clip}
Alec Radford, Jong~Wook Kim, Chris Hallacy, Aditya Ramesh, Gabriel Goh, Sandhini Agarwal, Girish Sastry, Amanda Askell, Pamela Mishkin, Jack Clark, et~al. 2021.
\newblock Learning transferable visual models from natural language supervision.
\newblock In \emph{International conference on machine learning}, pages 8748--8763. PmLR.

\bibitem[{Ramesh et~al.(2021)Ramesh, Pavlov, Goh, Gray, Voss, Radford, Chen, and Sutskever}]{dalle}
Aditya Ramesh, Mikhail Pavlov, Gabriel Goh, Scott Gray, Chelsea Voss, Alec Radford, Mark Chen, and Ilya Sutskever. 2021.
\newblock Zero-shot text-to-image generation.
\newblock In \emph{International conference on machine learning}, pages 8821--8831. Pmlr.

\bibitem[{Rombach et~al.(2022)Rombach, Blattmann, Lorenz, Esser, and Ommer}]{stable_diffusion}
Robin Rombach, Andreas Blattmann, Dominik Lorenz, Patrick Esser, and Bj{\"o}rn Ommer. 2022.
\newblock High-resolution image synthesis with latent diffusion models.
\newblock In \emph{Proceedings of the IEEE/CVF conference on computer vision and pattern recognition}, pages 10684--10695.

\bibitem[{Schuhmann et~al.(2022)Schuhmann, Beaumont, Vencu, Gordon, Wightman, Cherti, Coombes, Katta, Mullis, Wortsman et~al.}]{laion_2}
Christoph Schuhmann, Romain Beaumont, Richard Vencu, Cade Gordon, Ross Wightman, Mehdi Cherti, Theo Coombes, Aarush Katta, Clayton Mullis, Mitchell Wortsman, et~al. 2022.
\newblock Laion-5b: An open large-scale dataset for training next generation image-text models.
\newblock \emph{Advances in neural information processing systems}, 35:25278--25294.

\bibitem[{Schuhmann et~al.(2021)Schuhmann, Vencu, Beaumont, Kaczmarczyk, Mullis, Katta, Coombes, Jitsev, and Komatsuzaki}]{laion_1}
Christoph Schuhmann, Richard Vencu, Romain Beaumont, Robert Kaczmarczyk, Clayton Mullis, Aarush Katta, Theo Coombes, Jenia Jitsev, and Aran Komatsuzaki. 2021.
\newblock Laion-400m: Open dataset of clip-filtered 400 million image-text pairs.
\newblock \emph{arXiv preprint arXiv:2111.02114}.

\bibitem[{Wang et~al.(2024)Wang, Spinelli, Wang, Bai, Qin, and Chen}]{_encoder_3}
Haofan Wang, Matteo Spinelli, Qixun Wang, Xu~Bai, Zekui Qin, and Anthony Chen. 2024.
\newblock Instantstyle: Free lunch towards style-preserving in text-to-image generation.
\newblock \emph{arXiv preprint arXiv:2404.02733}.

\bibitem[{Wang et~al.()Wang, Wang, Xie, Qi, Shan, Wang, and Luo}]{_encoder_4}
Zhouxia Wang, Xintao Wang, Liangbin Xie, Zhongang Qi, Ying Shan, Wenping Wang, and Ping Luo.
\newblock Styleadapter: A unified stylized image generation model without test-time fine-tuning.

\bibitem[{Wen et~al.(2023)Wen, Jain, Kirchenbauer, Goldblum, Geiping, and Goldstein}]{hardpromptmadeeasy}
Yuxin Wen, Neel Jain, John Kirchenbauer, Micah Goldblum, Jonas Geiping, and Tom Goldstein. 2023.
\newblock Hard prompts made easy: Gradient-based discrete optimization for prompt tuning and discovery.
\newblock \emph{Advances in Neural Information Processing Systems}, 36:51008--51025.

\bibitem[{Xia et~al.(2021)Xia, Huang, Duan, Zhang, Ji, Sui, Cui, Bharti, and Zhou}]{pretrained_2}
Qiaolin Xia, Haoyang Huang, Nan Duan, Dongdong Zhang, Lei Ji, Zhifang Sui, Edward Cui, Taroon Bharti, and Ming Zhou. 2021.
\newblock Xgpt: Cross-modal generative pre-training for image captioning.
\newblock In \emph{Natural Language Processing and Chinese Computing: 10th CCF International Conference, NLPCC 2021, Qingdao, China, October 13--17, 2021, Proceedings, Part I 10}, pages 786--797. Springer.

\bibitem[{Xie et~al.(2024)Xie, Chen, Chen, Cai, Tang, Lin, Zhang, Li, Zhu, Lu, and Han}]{sana}
Enze Xie, Junsong Chen, Junyu Chen, Han Cai, Haotian Tang, Yujun Lin, Zhekai Zhang, Muyang Li, Ligeng Zhu, Yao Lu, and Song Han. 2024.
\newblock \href {https://arxiv.org/abs/2410.10629} {Sana: Efficient high-resolution image synthesis with linear diffusion transformers}.
\newblock \emph{Preprint}, arXiv:2410.10629.

\bibitem[{Zhang et~al.(2024)Zhang, Zhang, Dong, Zang, and Wang}]{longclip}
Beichen Zhang, Pan Zhang, Xiaoyi Dong, Yuhang Zang, and Jiaqi Wang. 2024.
\newblock Long-clip: Unlocking the long-text capability of clip.
\newblock \emph{arXiv preprint arXiv:2403.15378}.

\end{thebibliography}

\appendix


\begin{table*}[!ht]
\centering
\resizebox{\textwidth}{!}{%
\begin{tabular}{lllll}
\hline
\textbf{Poem} & \textbf{Weighted Prompt 1} & \textbf{Weighted Prompt 2} & \textbf{Weighted Prompt 3} & \textbf{Weighted Prompt 4} \\ \hline
\begin{tabular}[c]{@{}l@{}}Little Jack Horner\\ Sat in a corner,\\ Eating his Christmas pie; Ce\\ He put in his thumb,\\ And he pulled out a plum,\\ And said, “What a good boy am I!"\end{tabular} & \begin{tabular}[c]{@{}l@{}}Little Jack Horner (boy:1.5)  \\ Sat in a (corner:1.5),  \\ Eating his (Christmas pie:1.6);  \\ He put in his (thumb:1.5),  \\ And he pulled out a (plum:1.6),  \\ And said, “What a good (boy:1.5) am I!”\end{tabular} & \begin{tabular}[c]{@{}l@{}}Little Jack Horner  \\ Sat in a (corner:1.2),  \\ Eating his (Christmas pie:1.5);  \\ He put in his (thumb:1.2),  \\ And he pulled out a (plum:1.3),  \\ And said, “What a good boy am I!”\end{tabular} & \begin{tabular}[c]{@{}l@{}}(Little:0.9) (Jack:1.5) (Horner:1.5)  \\ Sat in a (corner:1.7),  \\ (Eating:0.9) his (Christmas:1.6) (pie:1.6);  \\ He put in his (thumb:1.5),  \\ And he pulled out a (plum:1.7),  \\ And said, “What a good boy am I!”\end{tabular} & \begin{tabular}[c]{@{}l@{}}Little Jack (Horner:1.5)  \\ Sat in a (corner:1.4),  \\ Eating his (Christmas:1.3) (pie:1.2);  \\ He put in his (thumb:1.1),  \\ And he pulled out a (plum:1.6),  \\ And said, “What a (good:0.8) (boy:0.9) am I!”\end{tabular} \\ \hline
\begin{tabular}[c]{@{}l@{}}What sound was that?\\ I turn away, into the shaking room.\\ What was that sound that came in on the dark?\\ What is this maze of light it leaves us in?\\ What is this stance we take,\\ To turn away and then turn back?\\ What did we hear?\\ It was the breath \\ we took when we first met.\\ Listen. It is here.\end{tabular} & \begin{tabular}[c]{@{}l@{}}What (sound:1.6) was that?  \\ I turn away, into the (shaking:1.7) room.  \\ What was that (sound:1.6) that\\  came in on the (dark:1.5)?  \\ What is this (maze:1.5) of \\ (light:1.6) it leaves us in?  \\ What is this (stance:0.8) we take,  \\ To turn away and then turn back?  \\ What did we (hear:1.5)?  \\ It was the (breath:1.6) \\ we took when we first met.  \\ Listen. It is (here:1.5).\end{tabular} & \begin{tabular}[c]{@{}l@{}}What (sound:1.2) was that?  \\ I turn away, into the (shaking:1.1) (room:1.3).  \\ What was that (sound:1.2) that \\ came in on the (dark:1.1)?  \\ What is this (maze:1.2) of \\ (light:1.3) it leaves us in?  \\ What is this (stance:1.1) we take,  \\ To turn away and then turn back?  \\ What did we hear?  \\ It was the (breath:1.3) \\ we took when we first met.  \\ Listen. It is (here:1.2).\end{tabular} & \begin{tabular}[c]{@{}l@{}}What (sound:1.7) was that?  \\ I turn away, into the (shaking:1.5) room.  \\ What was that (sound:1.7) that \\ came in on the (dark:1.5)?  \\ What is this (maze:1.6) of \\ (light:1.6) it leaves us in?  \\ What is this (stance:0.9) we take,  \\ To turn away and then turn back?  \\ What did we hear?  \\ It was the (breath:1.5) \\ we took when we first met.  \\ Listen. It is here.\end{tabular} & \begin{tabular}[c]{@{}l@{}}What (sound:1.8) was that?  \\ I turn away, into the (shaking:1.5) (room:1.3).  \\ What was that (sound:1.8) \\ that came in on the (dark:1.4)?  \\ What is this (maze:1.6) of\\  (light:1.5) it leaves us in?  \\ What is this (stance:1.2) we take,  \\ To turn away and then turn back?  \\ What did we hear?  \\ It was the (breath:1.9)\\  we took when we first (met:1.4).  \\ (Listen:1.1). It is here.\end{tabular} \\ \hline
\end{tabular}%
}
\caption{These are the original poems that are passed as an input to the diffusion model for the results demonstrated in Figure \ref{qualitative}.}
\label{qualitative_poems_results}
\end{table*}



\begin{figure*}[!htbp]
\centerline{\includegraphics[width=\textwidth]{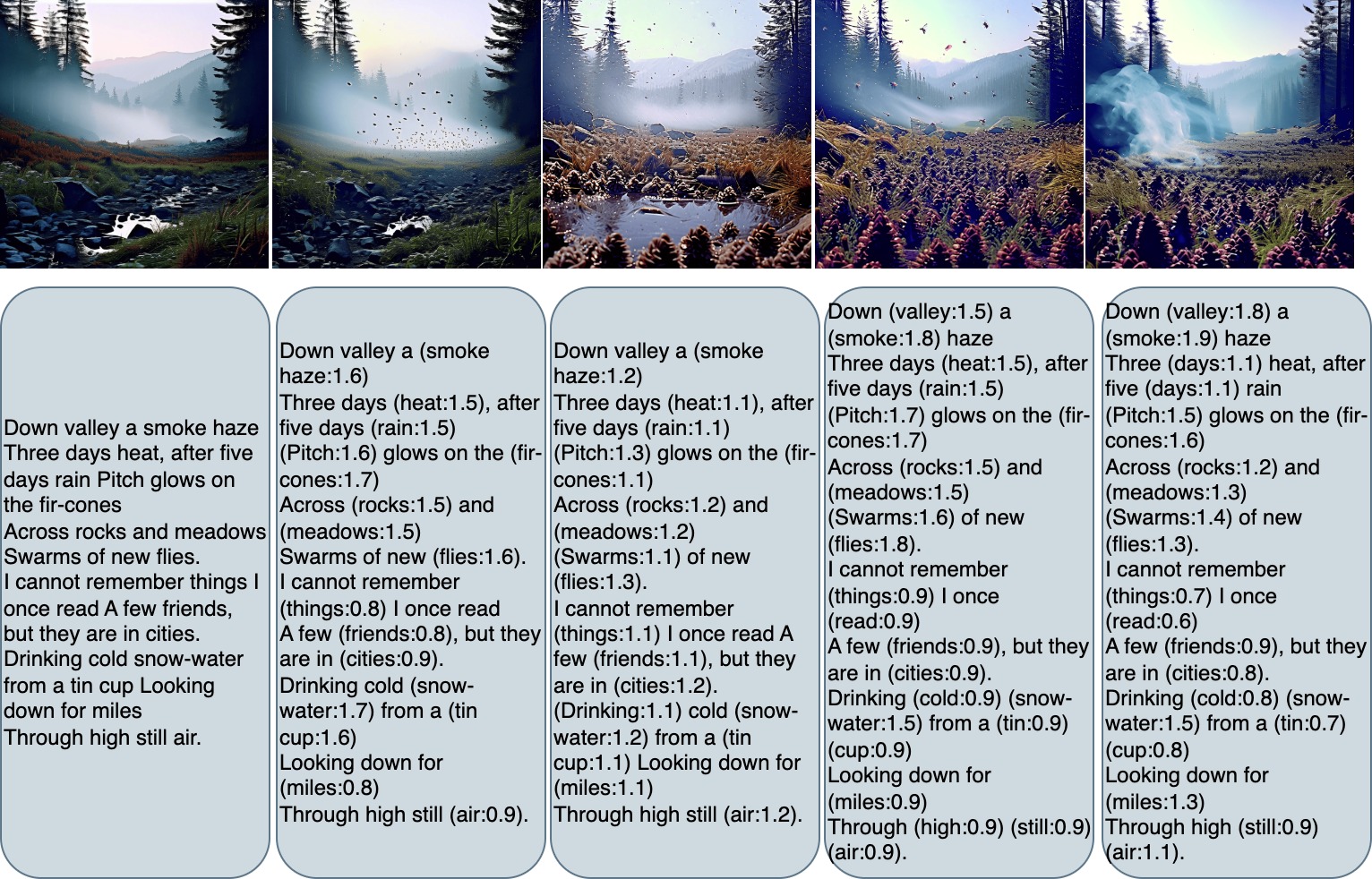}}
\caption{A comparison of generated images using different weights for various words in the same poem. All poems, along with their corresponding weighted prompts are provided in the Grey Box below.}
\label{qualitative2}
\end{figure*}

\section{List of Prompts}
\label{prompts}

This section presents the complete set of prompts we provided to GPT for the purpose of generating weighted prompts tailored to each poem. These prompts are specifically designed to emphasize particular words or phrases within the poetic text, such as key metaphors, emotionally rich expressions, or visually evocative language to guide the image generation model (e.g., Stable Diffusion) in focusing more on those elements. By strategically assigning higher importance to selected terms, we ensure that the resulting visual outputs more accurately reflect the intended artistic and semantic essence of the original poem.

\begin{table*}[!htbp]
\centering
\resizebox{\textwidth}{!}{%
\begin{tabular}{lll}
\hline
\textbf{Prompt 1:} & \textit{\begin{tabular}[c]{@{}l@{}}Refine the following poem into a weighted text prompt for text-to-image models. \\ Only apply weights to the most important visual words. Follow these strict rules:\\ \\ Identify and emphasize only the most critical visual elements. Avoid modifying too many words.\\ Use weight (1.5-1.8) for words that should be prominent in the generated image.\\ Use weight (0.7-0.9) for words that should appear less prominently.\\ Do not modify auxiliary, abstract, or transition words.\\ Maintain the structure and wording of the original poem.\\ Your response should only contain the weighted poem.\\ Example Input:\\ 'Underneath my outside face\\ There's a face that none can see.\\ A little less smiley,\\ A little less sure,\\ But a whole lot more like me.'\\ \\ Example Output:\\ 'Underneath my (outside:1.7) (face:1.7)\\ There's a (face:1.7) that none can see.\\ A little less (smiley:0.9),\\ A little less sure,\\ But a whole lot more like me.'\\ \\ Now apply these rules to the following poem:\end{tabular}} &  \\ \cline{1-2}
\textbf{Prompt 2:} & \textit{\begin{tabular}[c]{@{}l@{}}Prompt: Transform the following poem into a weighted text prompt for text-to-image generation. \\ Apply weights only to the most critical visual elements while preserving the poetic essence. \\ Follow these strict rules:\\ \\ Weighting Guidelines:\\ Incremental Weights (1.5 - 1.8) → Words that define the poem’s core visual or emotional identity\\ \\ Apply to words that strongly shape the imagery, mood, or metaphor.\\ Example: If the poem speaks of a storm, shadow, or teardrop, these evoke vivid visual elements and deserve higher weight.\\ Prioritize nouns (objects, scenery, emotions with physical manifestations).\\ Decremental Weights (0.7 - 0.9) → Words that modify or soften key visuals, but should not dominate\\ \\ Apply to words that exist only to describe or refine an image, rather than being the main focus.\\ Example: If a poem describes a smiley face but the mood suggests hidden sorrow, "smiley" should be weighted lower to reduce its dominance.\\ Use for adjectives or modifiers that subtly influence meaning but do not need strong emphasis.\\ DO NOT modify auxiliary words, transition words, or abstract concepts that lack direct visual impact (e.g., "that," "none," "sure," "because").\\ \\ Output Format:\\ Maintain the original poem’s structure.\\ Return only the transformed poem, with weights applied selectively and meaningfully.\\ Do not add explanations, notes, or comments.\\ Example Input:\\ 'Underneath my outside face\\ There's a face that none can see.\\ A little less smiley,\\ A little less sure,\\ But a whole lot more like me.'\\ \\ Example Output:\\ 'Underneath my (outside:1.7) (face:1.7)\\ There's a (face:1.7) that none can see.\\ A little less (smiley:0.9),\\ A little less sure,\\ But a whole lot more like me.'\\ \\ Now apply these rules to the following poem:\end{tabular}} &  \\ \cline{1-2}
\textbf{Prompt 3:} & \textit{\begin{tabular}[c]{@{}l@{}}Prompt: Refine the following poem into a weighted text prompt for text-to-image models. \\ Only apply weights to the most important visual words.\\ Your response should only contain the weighted poem.\end{tabular}} &  \\ \cline{1-2}
\textbf{Prompt 4:} & \textit{\begin{tabular}[c]{@{}l@{}}Prompt: Refine the following poem into a weighted text prompt for text-to-image models. \\ Only apply weights to the most important visual words. Follow these strict rules:\\ \\ Identify and emphasize only the most critical visual elements. Avoid modifying too many words.\\ Use weight (1.5-1.8) for words that should be prominent in the generated image.\\ Use weight (0.7-0.9) for words that should appear less prominently.\\ Do not modify auxiliary, abstract, or transition words.\\ Maintain the structure and wording of the original poem.\\ Your response should only contain the weighted poem.\end{tabular}} &  \\ \hline
\end{tabular}%
}
\caption{List of prompts used in our study.}
\label{promptstable}
\end{table*}

\begin{algorithm*}[!ht]
\scriptsize\centering
\begin{tabular}{cp{0.1in}c}
\begin{lstlisting}
Function parse_prompt_attention(text):
    Initialize regex patterns for parsing attention markers
    Initialize lists: result list (res), round_brackets, square_brackets
    Define multipliers: round_bracket_multiplier = 1.1, square_bracket_multiplier = 1/1.1

    Function multiply_range(start_position, multiplier):
        Multiply weights of all text segments from start_position to end

    Iterate over tokens extracted using regex:
        If token starts with "\" (escaped character), append it to result with weight 1.0
        If token is "(", push current position to round_brackets
        If token is "[", push current position to square_brackets
        If token is a closing bracket with weight, apply weight to last opened bracket
        If token is ")", apply round_bracket_multiplier to last opened round bracket
        If token is "]", apply square_bracket_multiplier to last opened square bracket
        Otherwise, split text on "BREAK" and append each segment with weight 1.0

    Apply multipliers to any unclosed brackets
    Merge adjacent segments with the same weight
    Return result list (text segments with weights)

Function get_prompts_tokens_with_weights(clip_tokenizer, prompt):
    Parse text into segments with weights using parse_prompt_attention()
    Initialize lists: text_tokens and text_weights

    For each (word, weight) in parsed text:
        Tokenize the word using CLIP tokenizer
        Append tokenized output to text_tokens
        Expand weight across tokens and append to text_weights

    Return text_tokens, text_weights

Function group_tokens_and_weights(token_ids, weights, pad_last_block=False):
    Define special tokens: bos (beginning of sentence), eos (end of sentence)
    Initialize 2D lists: new_token_ids, new_weights

    While token_ids list has 75 or more tokens:
        Extract first 75 tokens and weights
        Append them to new_token_ids and new_weights with bos and eos

    If remaining tokens exist:
        If pad_last_block is True, pad to 75 tokens with eos
        Append remaining tokens and weights

    Return new_token_ids, new_weights

Function get_weighted_text_embeddings_sdxl(pipe, prompt, prompt_2, neg_prompt, neg_prompt_2, num_images_per_prompt, device, clip_skip, lora_scale):
    Set device for computation
    Adjust LoRA scaling if necessary

    If prompt_2 exists, concatenate with prompt
    If neg_prompt_2 exists, concatenate with neg_prompt

    Convert prompts using textual inversion if applicable
    Define eos token

    Tokenize and extract weights for:
        - prompt_t1, neg_prompt_t1 (main tokenizer)
        - prompt_t2, neg_prompt_t2 (secondary tokenizer)

    Pad shorter prompt/negative prompt sets to match token length
    Initialize lists: embeds, neg_embeds

    Group tokens and weights into blocks of 77 tokens
    Iterate over each token block:
        - Convert tokens to tensor
        - Pass through text encoders
        - Extract embeddings
        - Apply weighting adjustments for emphasis
        - Append processed embeddings to embeds and neg_embeds lists

    Stack embeddings for final output
    Return prompt_embeds, neg_prompt_embeds
\end{lstlisting}
\end{tabular}
\caption{Entire Pseudocode for Weighted Prompt Manipulation (WPM) }
\label{wpmalgorithm}
\end{algorithm*}

\end{document}